%% file: ijcai24.tex
\documentclass{article}
\usepackage{ijcai24}

\usepackage{times}
\usepackage{soul}
\usepackage{url}
\usepackage[hidelinks]{hyperref}
\usepackage[utf8]{inputenc}
\usepackage[small]{caption}
\usepackage{graphicx}
\usepackage{amsmath}
\usepackage{amsthm}
\usepackage{booktabs}
\usepackage{algorithm}
\usepackage{algpseudocode}
\usepackage[switch]{lineno}

\title{Only My Model On My Data: A Privacy Preserving Approach Protecting one Model and Deceiving Unauthorized Black-Box Models}

\author{
Weiheng Chai$^*$
\and
Brian Testa$^*$\and
Huantao Ren\and
Asif Salekin\and
Senem Velipasalar\\
\affiliations
Syracuse University
\emails
\{wchai01, bptesta, hren11, asalekin, svelipas\}@syr.edu,
}

\begin{document}

\maketitle

\begin{abstract}
Deep neural networks are extensively applied to real-world tasks, such as face recognition and medical image classification, where privacy and data protection are critical. Image data, if not protected, can be exploited to infer personal or contextual information. Existing privacy preservation methods, like encryption, generate perturbed images that are unrecognizable to even humans. Adversarial attack approaches prohibit automated inference even for authorized stakeholders, limiting practical incentives for commercial and widespread adaptation. This pioneering study tackles an unexplored practical privacy preservation use case by generating human-perceivable images that maintain accurate inference by an authorized model while evading other unauthorized black-box models of similar or dissimilar objectives, and addresses the previous research gaps.
We validate the efficacy of our proposed solutions across three distinct datasets and diverse models.
The datasets employed are ImageNet, for image classification, Celeba-HQ dataset, for identity classification, and AffectNet, for emotion classification. Our results show that the generated images can successfully maintain the accuracy of a protected model and degrade the average accuracy of the unauthorized black-box models to 11.97\%, 6.63\%, and 55.51\% on ImageNet, Celeba-HQ, and AffectNet datasets, respectively.
\end{abstract}
\input{sec/1_intro}

\input{sec/2_relatedworks}

\input{sec/2a_attackmodel}
\input{sec/3_method}
\input{sec/4_experiments}
\input{sec/5_cross_task}
\input{sec/6_analysis}
\input{sec/7_ablationstudy}
\input{sec/8_conclusion}
\bibliographystyle{named}
\bibliography{ijcai24}

\end{document}


\maketitle

\section{Analysis of Feature-based Approach on SVHN and LFW datasets}
In this section, we analyze the efficacy of feature-based attacks on street number classification and face recognition tasks. The corresponding results are summarized in Tables~\ref{tab:alb_feature_svhn} and~\ref{tab:alb_feature_lfw}. In Table~\ref{tab:alb_feature_svhn}, we utilize VGG11 as a surrogate model to assess the robustness of both VGG16 and ResNet18. In Table~\ref{tab:alb_feature_lfw}, we initially set MobilefaceNet as the protected model with FaceNet as the surrogate, and then reverse their roles in the subsequent row. The observed outcomes suggest a lower effectiveness of feature-based attacks on these datasets compared to more complex tasks. This could be attributed to the relatively simpler class structure of these datasets; SVHN comprises only 10 classes, and face recognition can essentially be treated as a binary classification problem. These findings corroborate the conclusions drawn in Section.6.2, emphasizing that feature-based attacks are less effective for simpler classification tasks.

\begin{table}[h]
\resizebox{1\columnwidth}{!}{
\begin{tabular}{|c|c|c|c|c|c|}
\hline
Accuracy(\%) & VGG11 & VGG16 & ResNet18 & ResNet34 & Resnet50 \\ \hline
VGG11        & 100   & 99.7  & 98.7     & 99.1     & 99.55    \\ \hline
VGG16        & 99.5  & 100   & 97.6     & 99       & 99.4     \\ \hline
ResNet18     & 92.6  & 94.5  & 100      & 95.2     & 95.45    \\ \hline
ResNet34     & 92.35 & 94    & 90.5     & 100      & 94.6     \\ \hline
\end{tabular}}
\caption{Feature Map Distortion Results on SVHN dataset.}
\label{tab:alb_feature_svhn}
\end{table}

\begin{table}[h]
\centering
\resizebox{0.85\columnwidth}{!}{
\begin{tabular}{|c|c|c|c|}
\hline
          & Mobilefacenet & FaceNet & Spherefacenet \\ \hline
Mobilefacenet & 100       & 98.6    & 95        \\ \hline
FaceNet   & 100       & 100     & 99.73     \\ \hline
\end{tabular}}
\caption{Feature Map Distortion Results on LFW dataset.}
\label{tab:alb_feature_lfw}
\end{table}
\section{Image Quality}
\begin{figure*}
    \centering\resizebox{1.5\columnwidth}{!}{
    \includegraphics{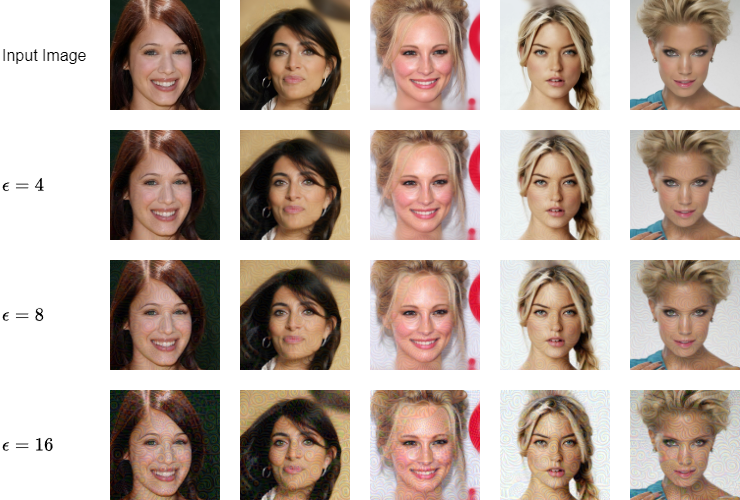}}

    \caption{Example generated images for different $\epsilon$ values.}
    \label{fig:quality}
\end{figure*}

Here we show more examples for images with different $\epsilon$ in Figure.~\ref{fig:quality}

%% file: sec/1_intro.tex
\section{Introduction}
\label{sec:intro}
In the contemporary landscape, the ubiquitous exchange and sharing of images plays an integral role across diverse platforms. These images serve various purposes, from the presentation of personal photos on LinkedIn profiles to institutional web pages and the verification process of ID cards.~Moreover, the digital realm witnesses the widespread sharing of images, contributing to various practical applications, such as surveillance systems, object recognition technologies, and the identification of license plates.
\par 
The essence of this image-sharing paradigm is deeply entrenched in the operational functionality of platforms, particularly in enabling the presentation and validation of crucial information related to users, employees, candidates, and more. This validation process incorporates a dual approach, \emph{employing visual human assessment} alongside \emph{AI-driven automated verification systems}.
\par 
Despite its indispensable utility, the act of sharing images triggers substantial privacy apprehension among users \cite{xu2011information,trepte2021social,de2020contextualizing}. 
Image data can be harvested to infer personal information about the subjects \cite{fei2020deep,wang2018deep,wang2022presentation,police}, 
or can be used to gather contextual information about the subject's surroundings \cite{vlontzos2019multiple,li2021framework}.

This paper embarks on a pioneering exploration aimed at mitigating these privacy concerns. The main goal is to provide a framework through which companies and institutions can harness user images for automated inference and human visualization while upholding user privacy. 
\par 
Towards the aforementioned goal, this paper addresses a fundamental question:~\textit{Can we strategically add minimal noise to an image for a specific task while satisfying three constraints: (i) ensuring these images maintain their perceptual functionality to humans; (ii) enabling a designated/ authorized model for a target task, trained by the stakeholder, to derive accurate inferences from these altered images, and (iii) causing any (black-box) unauthorized machine learning models to be incapable of generating accurate inferences for the same altered image in that same or even a different task?} 

We can broadly identify two lines of related research in this area: (i) On one side, there exists privacy-preserving methods that focus on image encryption \cite{huang2019toward} to prevent access by unintended 3rd-parties in such a way that the images become imperceptible by humans.~This makes these approaches not suitable for the context mentioned above, where one of the goals is to disseminate the images widely for human visual comprehension as well; (ii) the second line of research focuses on adversarial evasion techniques~\cite{chakraborty2018adversarial,carlini2017towards}, which aim to keep images still perceivable by humans (by restricting perturbations according to an L-norm constraint) while deceiving the inference capability of black-box machine learning models. These methods only focus on attacks and do not address the protection of an authorized model. 

That said, multiple works by Kwon et al.~\cite{kwon2018friend,kwon2019restricted,kwon2022priority} have considered this problem as a multi-objective loss function that allows development of adversarially evasive samples that both decrease the loss for an \textbf{authorized model} (for which we want inferences to be successful) while also increasing the loss for \textbf{unauthorized models} (which we want to degrade).~Yet, this approach requires a white-box setting, i.e. when generating image variants it is assumed that they have a priori knowledge of \textit{both} the authorized \textit{and} unauthorized models. 
\begin{figure}
    \centering\resizebox{1\columnwidth}{!}{
    \includegraphics{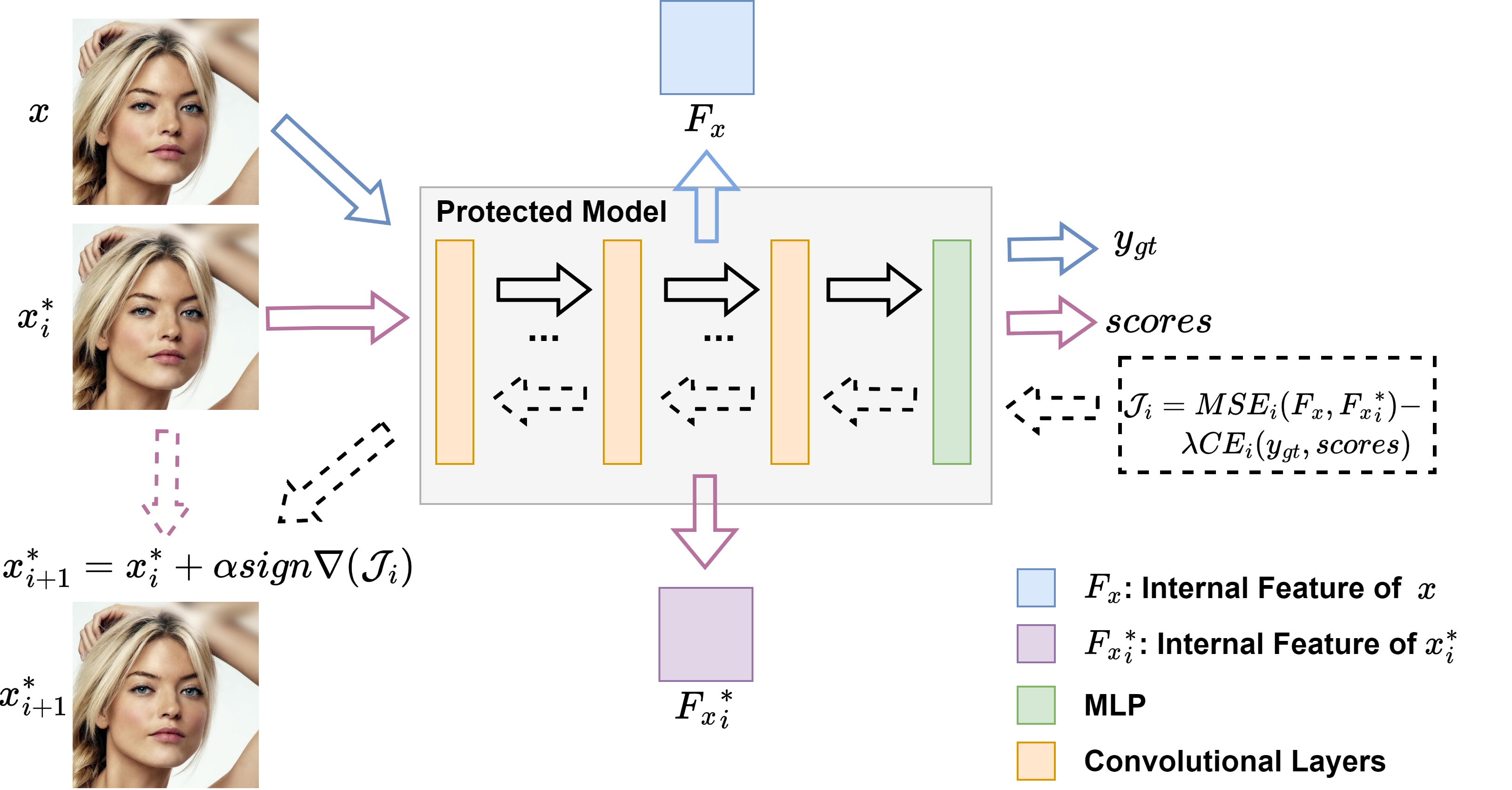}}
    \caption{Proposed feature-based privacy protection. $x$ denotes the input image, $x^*_i$ and $x^*_{i+1}$ denote the generated image after $i^{th}$ and  $i+1^{th}$ iteration, respectively. 
   }
    \label{fig:method}
\end{figure}

Our work is different from all the aforementioned approaches, and fills a gap not addressed by prior studies.~\textit{This work is the first of its kind to protect the privacy of image data by protecting a known, authorized network, while simultaneously degrading unknown, unapproved inference of other networks, and keeping the images still perceivable by humans}. 

Our approach works with a black-box setting for the unauthorized models. It is based on feature map distortion and does not require a surrogate model and assumes knowledge of only the authorized model (see Fig.~\ref{fig:method}). 
To accomplish this, the proposed approach relies on a key insight: early layers within a neural network perform extraction of features, which are then used for inference in later layers \cite{wang2020cnn,dosovitskiy2014discriminative,kim2017convolutional}. This work hypothesizes that disruptions to the feature representation would transfer to the feature extraction layers of other networks. So by optimizing for a multi-objective loss that both disrupts the feature representation \textbf{and} protects the inference performed by the authorized classifier, this should generate transferable adversarial examples (AEs) for previously unseen networks. 
Evaluations have been performed on multiple image datasets with various authorized and unauthorized classifiers.
As the pioneer paper to tackle the aforementioned practical objective, this work makes the following contributions:
\begin{itemize}
    \item We propose a novel method for data/privacy protection in different classification tasks, especially effective for tasks involving a relatively large number of classes.
    \item While protecting an authorized model, the proposed approach can successfully deceive unauthorized black-box models.
    \item We conduct experiments on various datasets to verify the effectiveness of our  approach. The first dataset is ImageNet~\cite{deng2009imagenet}, which is for general image classification. The other two datasets involve more privacy-sensitive aspects. More specifically, CelebA-HQ~\cite{liu2015celebahq} and AffectNet~\cite{mollahosseini2017affectnet} datasets are used for identity classification and facial expression classification task, respectively.
    \item We explore the feasibility of our approach when attacked models are designed for different tasks than the protected model, i.e. a cross-task scenario.
    \item We perform extensive ablation studies on different aspects of the proposed approach. 
\end{itemize}

%% file: sec/2_relatedworks.tex
\section{Related Work}
\label{sec:related}
There is a myriad of works that consider evasion of image classifiers, including both white-box \cite{goodfellow2014explaining,chakraborty2018adversarial} and black-box \cite{papernot2017practical,carlini2017towards} attacks. In a white-box scenario, the attacker possesses complete knowledge of the target model, including its architecture, parameters, weights, and even the training data. This enables the attacker to customize the attack, often using gradient-based methods to find the most effective perturbations to the input data.

In the black-box scenario~\cite{guo2019simple,ilyas2018black}, the attacker has limited or no knowledge of the target model.
Query-based methods were proposed, assuming the attackers' access to the input/output of the target model \cite{dang2017evading,juuti2019making}. Unlike query-based evasion, transferable attacks do not require querying and generate evasive samples across different models \cite{maho2021surfree}. 

In both cases, these attacks consider attack effectiveness, and are constrained by some notion of perceptability. This often comes in the form of maximum pixel perturbation $\epsilon$ to limit the amount of disruption caused by the attack. Yet, the goal is  designing very effective attacks, without focusing on protecting an authorized model. 

Kwon et al.~\cite{kwon2018friend,kwon2019restricted,kwon2022priority} addressed this shortfall by protecting a single image classifier and simultaneously disrupting one or more target image classifiers. These works have also been extended to other applications\cite{kwon2023toward,kwon2019selective,ko2023multi,kwon2021dual}. The image-based attack papers used \emph{the white-box}, transformation-based Carlini-Wagner attack~\cite{carlini2017towards}, and modified the loss function of the transformation network to include \textit{distortion}, \textit{friend}, and \textit{enemy} loss terms. The \textit{distortion} term penalized large changes to the original image (constraining the attack to some $\epsilon$), while the \textit{friend} and \textit{enemy} terms discouraged changes that impacted the protected (friend) classifier while degrading the targeted (enemy) classifiers. 

While achieving success in the intended objective, these studies required that the protected and targeted classifiers share the same topology. Moreover, they necessitated white-box access to both the protected and targeted classifiers. This white-box prerequisite significantly restricts their practicality and renders them inapplicable to the use cases this paper addresses.

%% file: sec/2a_attackmodel.tex
\section{Protect \& Attack Model}
\label{sec:attack}
We first provide the context under which this work was evaluated, including the attack goal, access levels and assumptions.

\textbf{Attack Goal.} Maintain inference performance of an authorized, white-box model while simultaneously degrading black-box, unauthorized models \textit{and} maintaining perceptual functionality for humans.


\textbf{Proposed System Access Level.} The \textbf{authorized model} is an image classifier for which we have access to the topology, weights, and inference results.~The \textbf{unauthorized models} are unknown black-box models for which we have neither access to the weights nor to the topologies. Further, we do not have query access to the target models. 

\textbf{Assumptions.} The primary assumption is that all of the models perform inference based upon image pixel data, and not based on any other format or metadata-specific information. This work further assumes that, before an image is shared with a broader group, the image can be preprocessed to introduce privacy-preserving perturbations. This assumes adequate access to the authorized model.

\textbf{Outcome.} The desired outcome is a set of adversarially evasive images that are accurately classified by the authorized model, misclassified by the unauthorized models, and perceptually similar to the original images. 

%% file: sec/3_method.tex
\section{Methodology} \label{sec:method}
The primary objective of our study is to create perturbed images that safeguard an authorized model while diminishing the performance of unauthorized target models. This objective is formulated as follows:
\begin{equation}
    \begin{aligned}
        & x^* =  \underset{x^*}{\text{argmax}}\quad \min_{\forall t \ne p } \mathcal{L}_t\\
        & \text{subject to} \quad f_p(x^*) = f_p(x),
    \end{aligned}
    \label{eq:obj_0}
\end{equation}
where \( x^* \) represents the generated image, $f_p$ is the authorized model, and \( \mathcal{L}_t \) signifies the loss of the target models.~Optimizing the image to satisfy $f_p(x^*) = f_p(x)$  presents practical challenges. Therefore, we pivot our strategy to focus on minimizing the task-specific loss \( \mathcal{L}_p \) of the authorized model. This revised objective is delineated as follows:
\begin{equation}
\footnotesize
    x^* =  \underset{x^*}{\text{argmax}}\quad \min_{\forall t \ne p } \mathcal{L}_t \quad \underset{x^*}  {\text{argmin}}\quad \mathcal{L}_p.
    \label{eq:obj_1}
\end{equation}
In contrast to \cite{kwon2018friend}, our work considers targeting unauthorized models within a black-box setting, aligning more closely with real-world scenarios. We propose a generalized solution to this problem, as detailed in Eq.~(\ref{eq:sol}), where \( \mathcal{L}' \) denotes the loss that is transferable to target models. When utilizing only an authorized model for generation, this loss originates from the authorized model, as shown in Eq.~(\ref{eq:feature_loss}). 
\begin{equation}
\small
    x^* =  \underset{x^*}{\text{argmax}} \quad \mathcal{L}' \quad \underset{x^*}  {\text{argmin}} \quad \mathcal{L}_p .
    \label{eq:sol}
\end{equation}
This can also be written as:
\begin{equation}
\small
    x^* =  \underset{x^*}{\text{argmax}}\quad \mathcal{J},
    \label{eq:sol_rewrite}
\end{equation}
where $\mathcal{J} = \mathcal{L}' - \lambda\mathcal{L}_p$, and $\lambda$ is the parameter to control the direction of the gradient.

During image generation, we iteratively update the image in accordance with the loss direction, as outlined in Algorithm \ref{alg:generate_process}. To maintain the visual quality of the generated image, we employ a budget constraint \( \epsilon \), utilizing the \( l_{\infty} \) norm as a measure. The impact of \( \epsilon \) on both performance and image quality is extensively discussed in Sec.~\ref{sec:quality}.
\begin{algorithm}
\footnotesize
\caption{Image Privacy Protection}
\textbf{Input} $\mathcal{L}'$: transferable attack loss; $\mathcal{L}_p$: task-specific loss for the authorized model; $x$: input image; $x^*$: generated image; $N$: number of  iterations; $\alpha$: the step size for updating the image; $\epsilon$: the budget for modifying the pixels
\begin{algorithmic}[1] 
\Procedure{GenerateImage}{}
    \State $x^{*}_{0} = x, i = 1$
    \While{$i < N+1$}
        \State $\mathcal{J}_{i-1} \gets \mathcal{L}'(x^{*}_{i-1}) - \lambda\mathcal{L}_{p}(x^{*}_{i-1})$
        \State $x^{*}_{i} \gets x^{*}_{i-1} + \alpha \text{sign}(\nabla(\mathcal{J}_{i-1}))$
        \State $x^{*}_{i} \gets \min(x^{*}_{i}, x + \epsilon, 255)$
        \State $x^{*}_{i} \gets \max(x^{*}_{i}, x - \epsilon, 0)$
        \State $i \gets i+1$

    \EndWhile
    \State \textbf{return} $x^{*}_{N}$
\EndProcedure
\end{algorithmic}
\label{alg:generate_process}
\end{algorithm}
%

As stated previously, our objective is to evade unauthorized black-box classifiers while protecting an authorized white-box classifier. 
To achieve this goal, we propose a \textit{feature map distortion} (FMD) approach, which generates the perturbed image with only the authorized model by using the Mean squared Error (MSE) of the internal feature map as the transferable loss ($\mathcal{L}'$). The cross-entropy (CE) loss of the authorized model is used as the task-specific loss ($\mathcal{L}_p$). Thus, the loss $\mathcal{J}$ in Eq.~(\ref{eq:sol_rewrite}) can be expressed as follows:
\begin{equation}
\footnotesize
    \mathcal{J} = \text{MSE}(f_{feat}(x), f_{feat}(x^{*})) - \lambda\text{CE}(f_{logits}(x^{*}), y_{gt}),
    \label{eq:feature_loss}
\end{equation}
where $f_{feat}$ denotes the output of the internal convolution layers of the authorized model, $f_{logits}(x^{*})$ is the output score of the authorized model with image $x^{*}$, and $y_{gt}$ is the ground-truth label of the original image $x$. When initially generating $x^{*}_{0}$ as $x^{*}_{0} = x$, the gradient of the MSE loss will be 0. Thus, we uniformly randomly initialize $x^{*}_{0}$ in the $[x-\epsilon, x+\epsilon]$ range.
\par 
The proposed $\mathcal{J}$ loss in Eq.~(\ref{eq:feature_loss}) serves the dual objective: degrading the performance of unauthorized models while safeguarding the authorized one. First, employing the MSE loss creates a feature representation of the perturbed image $x^*$, modified by adding noise, which significantly differs from the original feature map. This variance leads to misclassification, meaning there is no valid input-to-output path for commonly trained models $\forall f_t(x^*)$ with the same objective. However, considering the nature of deep learning models that establish numerous paths from input to output \cite{wang2021convolutional}, the use of CE loss ensures the retention of at least one pathway from input to output for the authorized model $f_p(x^*)$, serving the second objective of safeguarding it.
\par 

%% file: sec/4_experiments.tex
\section{Experimental Results}
\label{sec:exp}
We commence by presenting results from our proposed protection approach, which only employs the authorized model to generate the perturbed images, as introduced in Sec.~\ref{sec:method}. This approach is applied to different classification tasks, with varying number of classes, using three different datasets, namely ImageNet~\cite{deng2009imagenet}, Celeba-HQ \cite{liu2015celebahq}, and Affectnet \cite{mollahosseini2017affectnet} datasets.


\subsection{Results on the ImageNet Dataset}
\label{sec:exp_imgnet}
We use six pretrained models from the torchvision~\cite{torchvision}, namely VGG11 \cite{simonyan2014vgg}, VGG16 \cite{simonyan2014vgg}, ResNet18 \cite{he2016resnet}, ResNet34 \cite{he2016resnet}, Wide-ResNet50-2 \cite{zagoruyko2016wide}, and MobileNet-V2 \cite{sandler2018mobilenetv2}, which have been trained on the ImageNet dataset, which includes 1000 classes.~From the ImageNet validation set, we selected 5,000 images,
which were correctly classified by all six models. We set the budget parameter \( \epsilon \) as 16, step size \( \alpha \) as 4, number of steps \( N \) as 100, and the weight factor \( \lambda \) as 1. The most effective layer was chosen for each model to obtain the results presented in Table \ref{tab:imgnet}. Detailed discussion on layer selection and parameter settings can be found in Sec.~\ref{sec:layer_select} and \ref{sec:parameter}, respectively. 

The models in the first column of Table \ref{tab:imgnet} are the authorized models used during image generation, while the models listed in the top row are the unauthorized models used for testing. Thus, diagonal entries indicate the accuracy of the authorized model. As evident from the table, we achieve a remarkable average protected accuracy of 100\% while significantly reducing the accuracy of the target models from 100\% to an average of just 11.97\% across six different models.


\begin{table}[hb!]
\resizebox{1\columnwidth}{!}{
\begin{tabular}{|c|c|c|c|c|c|c|}\hline
  Accuracy (\%)                & VGG11 & VGG16 & ResNet18 & ResNet34 & W-Res50-2 & Mob-v2 \\\hline
VGG11             & 100                       & 15.38                     & 10.56                        & 20.68                        & 12.66                                 & 14.26                             \\\hline
VGG16             & 6.62                      & 100                       & 8.54                         & 16.54                        & 9.82                                  & 9.98                              \\\hline
ResNet18          & 12.52                     & 8.78                      & 100                          & 25.5                         & 16.36                                 & 10.18                             \\\hline
ResNet34          & 8.72                      & 5.36                      & 5.78                         & 100                          & 12.46                                 & 6.78                              \\\hline
W-Res50-2 & 7.14                      & 5.32                      & 4.96                         & 14.92                        & 100                                   & 9.28                              \\\hline
Mob-v2     & 10.02                     & 10.4                      & 13.18                        & 26.02                        & 20.34                                 & 100                              \\\hline
\end{tabular}}
\vspace{-0.2cm}
\caption{Experimental results on the ImageNet dataset with six different models on a 1000-category classification task.}
\label{tab:imgnet}
\end{table}
\subsection{Results on Celeba-HQ dataset}
\label{sec:exp_celeba}
We employ the same models as in Sec.~\ref{sec:exp_imgnet}, fine-tuning them on the Celeba-HQ dataset containing 30,000 images to classify faces. We selected images of 307 subjects, each with 15 or more images. A split of 4263 images (80\%) for training and 1215 images (20\%) for testing was used. Post-training, we selected 634 testing images that were correctly classified by all six models for our experiments. The parameters \( \alpha \), \( \epsilon \), $N$, and \( \lambda \) were kept consistent with those used in Sec.~\ref{sec:exp_imgnet}.
%
%
Results are presented in Tab.~\ref{tab:celeba}. The models in the first column represent the authorized models, while the model names in the top row are the attacked models for testing. The diagonal entries indicate the protected accuracy. We achieve an average protected accuracy of 99.92\% while significantly reducing the average accuracy of target models from 100\% to just 6.63\% across six different models.
\begin{table}[h!]
\resizebox{1\columnwidth}{!}{
\begin{tabular}{|c|c|c|c|c|c|c|}\hline
Accuracy (\%)                & VGG11 & VGG16 & ResNet18 & ResNet34 & W-Res50-2 & Mob-v2 \\\hline
VGG11             & 100                       & 2.05                      & 5.99                         & 7.73                         & 9.78                                  & 3.79                              \\\hline
VGG16             & 2.37                      & 99.53                     & 4.73                         & 6.78                         & 7.73                                  & 1.1                               \\\hline
ResNet18          & 2.52                      & 1.1                       & 100                          & 5.84                         & 8.04                                  & 1.1                               \\\hline
ResNet34          & 3                         & 1.1                       & 3.79                         & 100                          & 12.78                                 & 1.1                               \\\hline
W-Res50-2 & 11.99                     & 2.68                      & 4.1                          & 5.52                         & 100                                   & 3.6                               \\\hline
Mob-v2     & 21.45                     & 10.09                     & 11.51                        & 13.41                        & 22.24                                 & 100        \\\hline                      
\end{tabular}}
\vspace{-0.2cm}
\caption{Experimental results on the Celeba-HQ dataset on a 307-category classification task.}
\label{tab:celeba}
\end{table}
\subsection{Results on the AffectNet dataset}\label{sec:exp_affectnet}
We employ the same models as in Sec.~\ref{sec:exp_imgnet}, fine-tuning them for facial expression classification on the AffectNet dataset. AffectNet contains eight facial expressions: neutral, happy, angry, sad, fear, surprise, disgust, and contempt. The owner splits the data, such that 287,651 images are for training and 3,999 for testing. Post-training, we selected 1346 testing images that were correctly classified by all six models for our experiments. The parameters \( \alpha \), \( \epsilon \), and $N$ were kept consistent with those used in Sec.~ \ref{sec:exp_imgnet}. \( \lambda \) setting is determined by experiments.
The results are presented in Tab.~\ref{tab:affectnet}. The models in the first column represent the authorized models used during image generation, while the models in the top row are attacked models used for testing. The diagonal entries indicate the protected accuracy. We achieve an average protected accuracy of 99.67\%, while significantly reducing the average accuracy of target models from 100\% to just 55.5\% across six different models. The performance drop on the unauthorized models for this task is not as significant as that achieved with the classification tasks on the other two datasets. We deduce that this is related to this task being less complex than others. 
More specifically, in Sec.~\ref{sec:exp_imgnet} and~\ref{sec:exp_celeba}, the classification tasks involve 1000 and 307 classes, respectively. With AffectNet, on the other hand, there are only eight classes. Our hypothesis is further verified in Sec.~\ref{sec:task_comp}.

\begin{table}[hb!]
\resizebox{1\columnwidth}{!}{
\begin{tabular}{|c|c|c|c|c|c|c|}\hline
Accuracy (\%)                & VGG11 & VGG16 & ResNet18 & ResNet34 & W-Res50-2 & Mob-v2 \\\hline
VGG11         & 99.78                     & 47.92                     & 57.88                        & 66.57                        & 70.13                           & 58.47                       \\\hline
VGG16         & 55.72                     & 99.48                     & 62.85                        & 70.06                        & 75.71                           & 70.58                       \\\hline
ResNet18      & 51.19                     & 44.28                     & 100.00                       & 42.94                        & 51.19                           & 50.67                       \\\hline
ResNet34      & 61.74                     & 42.05                     & 26.30                        & 98.74                        & 60.55                           & 54.83                       \\\hline
W-Res50-2   & 42.27                     & 37.00                     & 35.88                        & 36.26                        & 100.00                          & 42.35                       \\\hline
Mob-v2       & 69.69                     & 60.10                     & 66.42                        & 78.01                        & 75.78                           & 100.00              \\\hline       
\end{tabular}}
\vspace{-0.2cm}
\caption{Experimental results on the AffectNet dataset for 8-class classification.}
~\label{tab:affectnet}
\vspace{-0.3cm}
\end{table}

%% file: sec/5_cross_task.tex
\vspace{-0.2cm}
\section{Cross-task feasibility study}
The results presented above in Sec.~\ref{sec:exp} show that the generated images from an authorized model can deceive unauthorized target models performing the same task. Since we employ the MSE loss between the internal feature maps as a transferable loss candidate, we further explore if our proposed approach can transfer across \textit{different tasks}. In other words, we performed additional experiments to study if the images generated from an authorized model designed for task A can also deceive unauthorized target models designed for task B, where A and B are different tasks.
Here, different tasks could involve many classes, e.g., image classification, object detection, etc., or a smaller number of classes, e.g., facial expression classification,  ethnicity classification, etc. 
\begin{table}[]
\resizebox{1\columnwidth}{!}{
\begin{tabular}{|c|c|c|c|c|c|c|c|}\hline
   mAP(\%)      & Original                  & VGG11 & VGG16 & ResNet18 & ResNet34 & W-Res50-2 & Mob-v2 \\\hline       
Faster-RCNN         & 100                        & 3.48                      & 2.63                      & 4.62                         & 4.02                         & 4.43                                  & 5.61                              \\\hline
RetinaNet          & 100                        & 3.49                      & 2.55                      & 4.94                         & 4.24                         & 4.38                                  & 6.26                              \\\hline
MobileNet-V3-SSD & 100                        & 2.71                      & 1.89                      & 5.58                         & 5.5                          & 3.18                                  & 3.79 \\\hline
\end{tabular}}
\vspace{-0.2cm}
\caption{The cross-task experiment results on the ImageNet dataset.}
\label{tab:fastrcnn}
\end{table}

\begin{figure}[h!]
    \centering\resizebox{0.7\columnwidth}{!}{
    \includegraphics{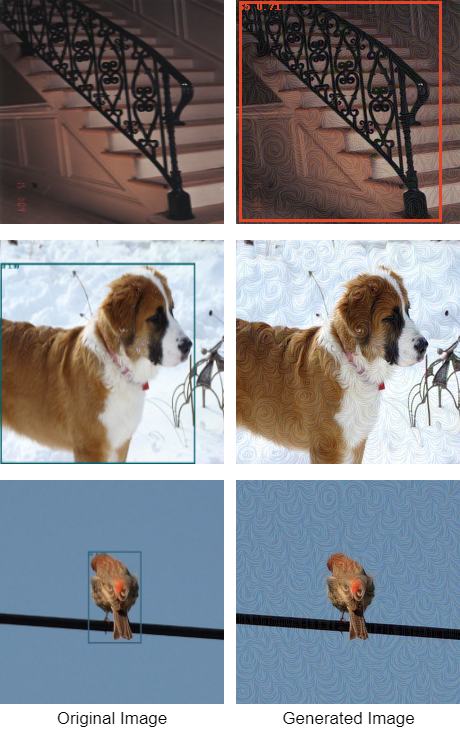}}
    \vspace{-0.3cm}
    \caption{Examples of object detection results from Mobile\_Net\_V3\_SSDLite with the original and generated images showing the false (first row) and missed detections (2nd and 3rd rows).}
    \label{fig:det}
\end{figure}

In the first cross-task experiment, we protect image classification models and attack object detection models. We generate the images from ImageNet, and attack three object detection models, namely Faster-RCNN~\cite{ren2015faster}, RetinaNet~\cite{lin2017focal}, and MobileNet-V3-SSDLite~\cite{liu2016ssd}, which are trained on the COCO dataset~\cite{lin2014microsoft}. In Tab.~\ref{tab:fastrcnn}, 
the models in the first row are the authorized models, from which the images are generated, and the model listed in the first column are the attacked object detection models. In this evaluation, we consider the output from the original image as the ground truth, so the original mean average precision (mAP) is 100\%. Then, we calculate the mAP on the generated images based on this ground truth. It can be seen that with our proposed method, when protecting six different models, the generated images can drop the average mAP of three target object detection models to 4.13\%. Different factors can cause this drop in mAP:
(i) bounding-boxes appear or disappear compared to the original image input; (ii) bounding-boxes shift compared to the original image; (iii) predicted class changes for a specific object compared with the original image. Some example detection results are shown in Fig.~\ref{fig:det}, where objects are completely missed in the 2nd and 3rd rows, and a false detection appears in the 1st row. 

\par
In the second cross-task experiment, we use two different datasets and generate images from the Celeba-HQ dataset and AffectNet datasets by protecting models trained for face identity and facial expression classification, respectively. We attack an open-source ethnicity classification model~\cite{serengil2020lightface} as the target model, which has been trained for a different task than the authorized models. The results are shown in Tab.~\ref{tab:ethnicity}.    
We consider the prediction of the original images as ground truth since we do not have the ground truth label for ethnicity, so the original accuracy is 100\%. It can be seen that with our proposed method, when protecting six different models, the generated images can drop the average accuracy to 74.39\% and 63.50\% on Celeba-HQ and AffectNet datasets, respectively.
\begin{table}[h!]
\resizebox{1\columnwidth}{!}{
\begin{tabular}{|c|c|c|c|c|c|c|c|}\hline
  Accuracy (\%)       & Original           & VGG11 & VGG16 & ResNet18 & ResNet34 & W-Res50-2 & Mob-v2 \\\hline
Celeba-HQ   &100 & 76.92                      & 76.92                      & 76.92                         & 73.08                         & 73.08                          & 69.23   \\\hline      
Affectnet &100 & 71.55                      & 72.51                      & 59.58                         & 59.73                         & 52.82                          & 64.78                      \\\hline
            
\end{tabular}}
\vspace{-0.2cm}
\caption{Experimental results on Celeba-HQ and Affectnet datasets for across task experiment.}
\label{tab:ethnicity}
\end{table}

Notably, all the experiments are done with the images generated from the corresponding datasets in Sec.~\ref{sec:exp}. It can be seen that for the ImageNet dataset when we protect image classification models, we can successfully deceive the unauthorized object detection models. For the evaluation when attacking an ethnicity classification model, the attack performance on the AffectNet dataset is good, with the average accuracy dropping to 63.50\%. On the Celeba-HQ dataset, the average accuracy only drops to 74.39\%. The results show that the cross-task transferability is affected by the tasks of the protected and attacked models, and by model characteristics. Future work will focus on designing a privacy protection method that can deceive unauthorized models, designed for different tasks, by reducing the dependency of the performance on the authorized model.

%% file: sec/6_analysis.tex
\section{Analysis of Results}
\label{sec:analysis}

The results above demonstrate that FMD is an effective method for evading previously unseen black-box models while protecting inference performed by a known authorized model. The intuition regarding this method is described briefly in Sec.~\ref{sec:intro} and \ref{sec:method}.~This section attempts to provide concrete analysis to support this intuition. 

\begin{figure}[t!]
    \centering\resizebox{0.85\columnwidth}{!}{
    \includegraphics{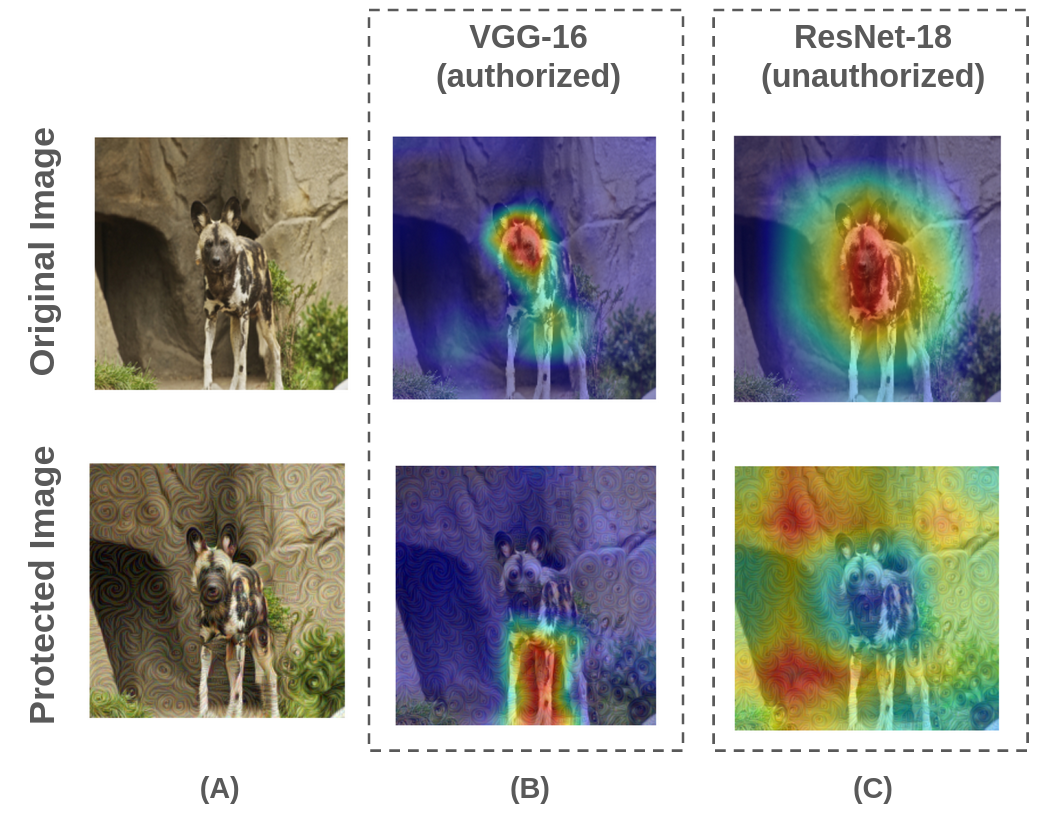}}
    \vspace{-0.4cm}
    \caption{(A) Example image and a protected variant generated by applying FMD on VGG16. (B) Grad-CAM heat maps for these images for VGG16. (C) Grad-CAM heat maps for ResNet18. }
    \label{fig:heat}
    \vspace{-0.2cm}
\end{figure}

Consider Fig.~\ref{fig:heat}, which shows an example image from the ImageNet dataset and a protected variant of that image generated by applying FMD to a VGG16 classifier. The heat maps, generated using Grad-CAM~\cite{selvaraju2016grad}, show the important areas of the original and perturbed images when considered by both VGG16 (the authorized classifier) and ResNet18 (an unauthorized classifier). We can see that for the authorized model, the area of interest shifts, whereas for the unauthorized model, the heat map is almost negated with a very diffused area of interest. 

\begin{figure}[t!]
    \centering\resizebox{0.85\columnwidth}{!}{
    \includegraphics{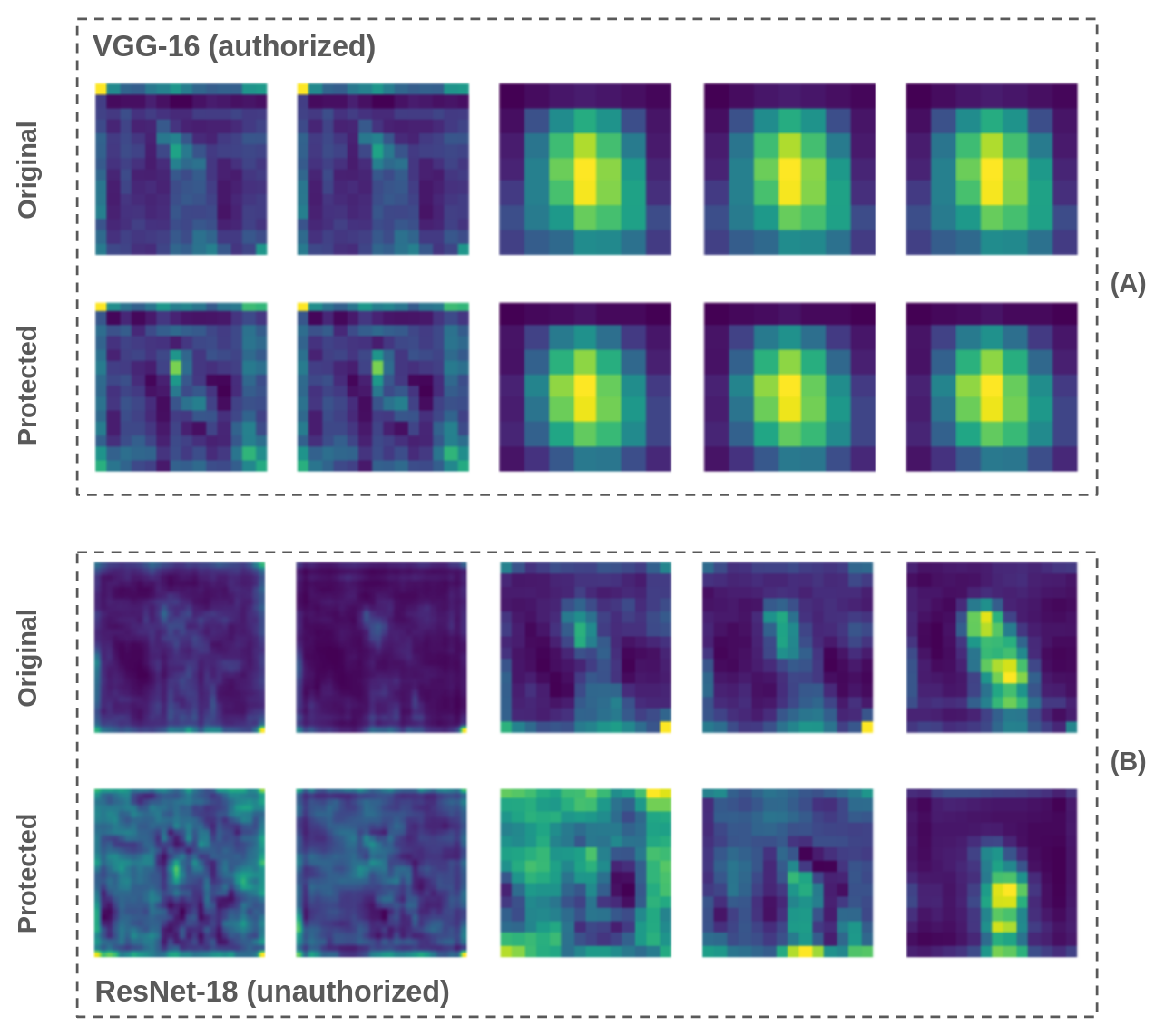}}
    \vspace{-0.2cm}
    \caption{(A) Final convolution layers in VGG16. (B) Final convolution layers in ResNet18. }
    \label{fig:heat2}
\end{figure}

Fig.\ref{fig:heat2} visualizes the features from the final Conv2d layers in both VGG16 and ResNet18 and shows two different scenarios within the classifiers. In Fig.~\ref{fig:heat2} (A) we see that the final convolution layers had very similar activations before the data was passed on to the inference layers at the end of the model. This makes sense since the authorized model correctly classified both the original and perturbed/protected images. In Fig.~\ref{fig:heat2} (B), on the other hand, we see a significant difference in these final feature maps, which translates into misclassifications by the unauthorized model.

\emph{Notably, these visualizations demonstrate the effectiveness of the $\mathcal{J}$ loss in Eq.~(\ref{eq:feature_loss}).} The MSE loss in Eq.~(\ref{eq:feature_loss}) successfully alters the original feature map as demonstrated in Fig.~\ref{fig:heat}, resulting in the absence of valid input-to-output path for black box unauthorized models, as shown in Fig.~\ref{fig:heat2} (B). Meanwhile, a path for the authorized model is retained, utilizing the cross-entropy (CE) loss, as demonstrated in Fig.~\ref{fig:heat2} (A).

\begin{figure}[t!]
    \centering\resizebox{0.45\columnwidth}{!}{
    \includegraphics{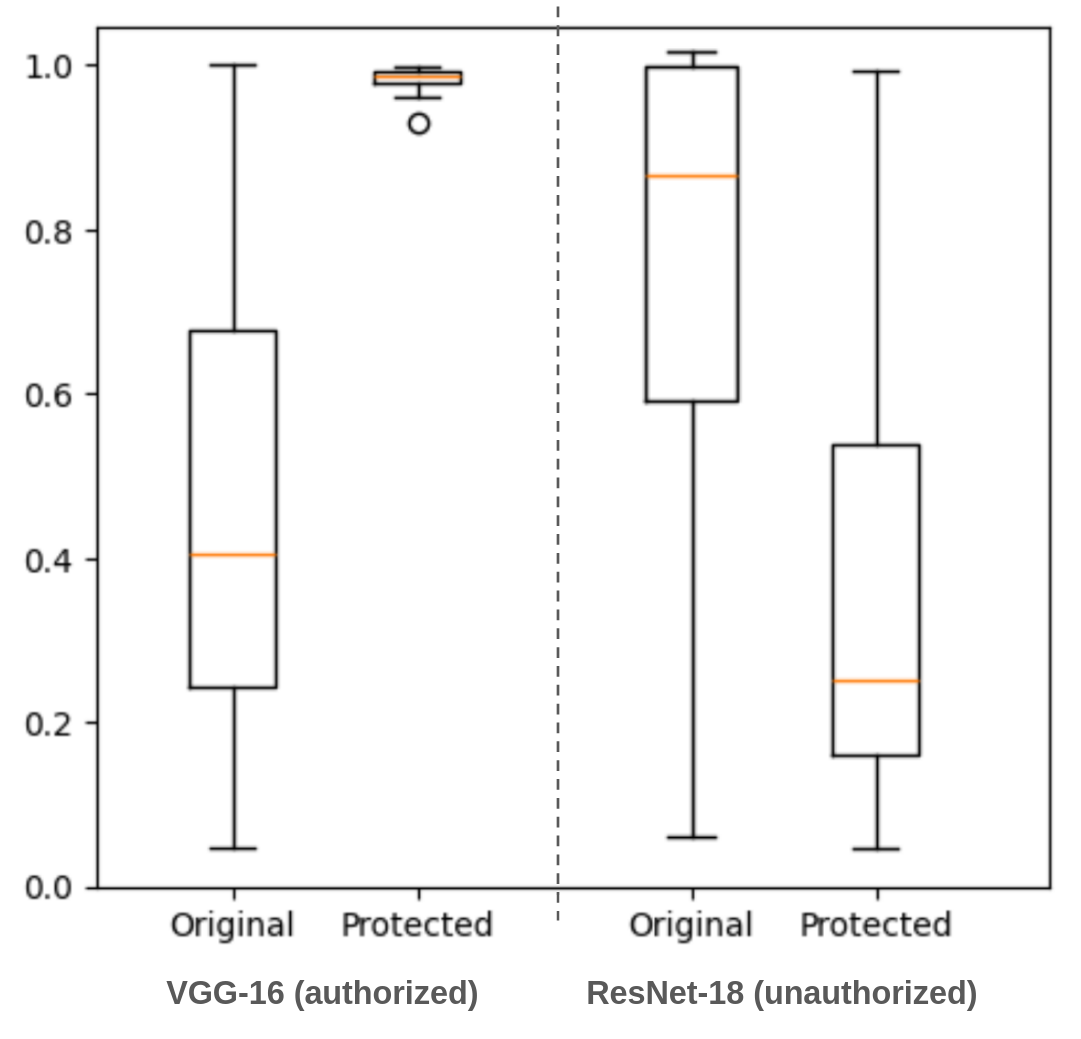}}
    \vspace{-0.2cm}
    \caption{Highest softmax values for the authorized and unauthorized models on the original and protected images. }
    \label{fig:softmax}
\end{figure}

We also investigated the impact of FMD-based protection on softmax outputs of models. In Fig.~\ref{fig:softmax}, we observed a notable increase in the largest softmax values for a sample of 1000 images from the ImageNet dataset and their protected variants generated using FMD on the VGG16 model. For VGG16, this resulted in significant overfitting to the model. Conversely, on the unauthorized model ResNet18, we observed a decrease in maximum softmax values, particularly for incorrect classes, along with a more homogeneous distribution of overall softmax values. While these values are not confidence scores, as the models were not calibrated for confidence representation \cite{guo2017calibration}, the indication is that FMD moves protected image variants away from the decision boundary for the authorized classifier.

%% file: sec/7_ablationstudy.tex
\section{Ablation Studies} \label{sec:abl}
%
\subsection{Layer Selection for FMD}\label{sec:layer_select}
In our FMD approach, we need to select a layer from the protected model to calculate $MSE(f_{feat}(x), f_{feat}(x^{*}))$. In our experiments, we randomly select 1000 images from the correctly classified images, and test all of the 13 convolution layers for protecting VGG16. The results in Tab.~\ref{tab:layer_selection} show that before layer 9, all the protection accuracy numbers are equal to or higher than 99.8\%. Layer 7 provides the best performance in degrading attacked models' performance. These trends are consistent across almost all the models. For the evaluation on MobileNet\_v2, the performance of target model is slightly lower for layer 6, but the difference is small.
\begin{table}[hb!]
\resizebox{1\columnwidth}{!}{
\begin{tabular}{|c|c|c|c|c|c|c|}\hline
Accuracy (\%)                & VGG11 & VGG16 & ResNet18 & ResNet34 & W-Res50-2 & Mob-v2 \\\hline
layer 0                    & 52.8                      & 99.8                      & 58.3                         & 68                           & 67.4                            & 51.3                        \\\hline
layer 1                    & 48.4                      & 100                       & 55.2                         & 64.1                         & 65.4                            & 47.2                        \\\hline
layer 2                    & 19.8                      & 99.8                      & 30.6                         & 41.1                         & 29.3                            & 12.9                        \\\hline
layer 3                    & 19                        & 99.8                      & 37.3                         & 45.1                         & 28.3                            & 12.9                        \\\hline
layer 4                    & 14                        & 99.8                      & 36.9                         & 45.7                         & 29.6                            & 15.7                        \\\hline
layer 5                    & 9.7                       & 99.8                      & 21.3                         & 32.1                         & 20.6                            & 13                          \\\hline
layer 6                    & 8                         & 99.8                      & 11.9                         & 20.8                         & 11.6                            & \textbf{8.9}                         \\\hline
layer 7                    & \textbf{8}                         & 99.8                      & \textbf{8.5}                          & \textbf{16.9}                         & \textbf{10.4}                            & 9.8                         \\\hline
layer 8                    & 22.2                      & 100                       & 27.1                         & 38.3                         & 26                              & 23.3                        \\\hline
layer 9                    & 36.3                      & 99.8                      & 50.2                         & 57.2                         & 40.8                            & 39.1                        \\\hline
layer 10                   & 31.5                      & 97.4                      & 49                           & 55.7                         & 42.7                            & 40.9                        \\\hline
layer 11                   & 39.6                      & 91.7                      & 53.2                         & 61                           & 48.3                            & 45.2                        \\\hline
layer 12                   & 44                        & 64                        & 55.8                         & 64.3                         & 51.8                            & 44   \\\hline                      
\end{tabular}}
\vspace{-0.2cm}
\caption{Results for layer selection when protecting VGG16}
\label{tab:layer_selection}
\end{table}
\subsection{Parameter Setting for FMD}\label{sec:parameter}
We have tested different $\alpha$ and $N$ values with the best-performing layer when protecting VGG16, using the same images used in Sec.~\ref{sec:layer_select}. We first fix $N$ to 100 and test different $\alpha$ settings. The results in Tab.~\ref{tab:alpha} show that when $\alpha=4$, the performance is slightly better than when $\alpha=8$. The accuracy can be affected by the random initialization described in Sec.~\ref{sec:method}, so both values of $\alpha = 4$ or $\alpha=8$ are acceptable. 
\begin{table}[h!]
\resizebox{1\columnwidth}{!}{
\begin{tabular}{|c|c|c|c|c|c|c|}\hline
Accuracy (\%)                & VGG11 & VGG16 & ResNet18 & ResNet34 & W-Res50-2 & Mob-v2 \\\hline
$\alpha=1$                    & 10.8                      & 100                       & 11.3                         & 21.9                         & 12.8                            & 12.4                        \\\hline
$\alpha=2$                    & 8.1                       & 99.8                      & 9.2                          & 19.9                         & 11.5                            & 10.8                        \\\hline
$\alpha=4$                    & 8                         & 99.8                      & 8.5                          & 16.9                         & 10.4                            & 9.8                         \\\hline
$\alpha=8$                    & 7.8                       & 99.8                      & 7.6                          & 16.2                         & 11.3                            & 10.9     \\\hline                  
\end{tabular}}
\vspace{-0.2cm}
\caption{Results for protecting VGG16 under different $\alpha$ settings.}
\label{tab:alpha}
\vspace{-0.2cm}
\end{table}

\begin{figure}[h!]
\vspace{-0.2cm}
    \centering\resizebox{0.8\columnwidth}{!}{
    \includegraphics{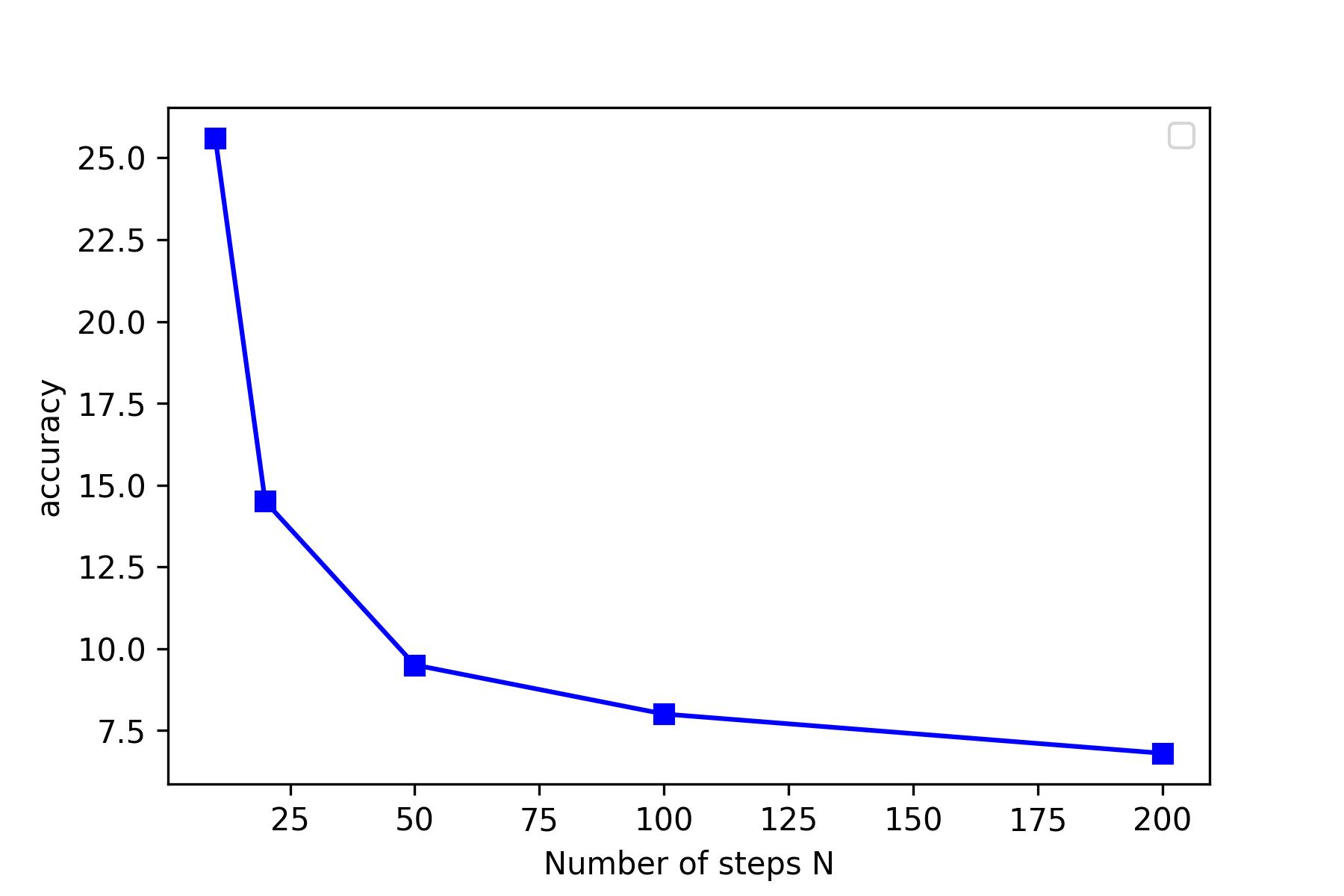}}
    \vspace{-0.3cm}
    \caption{Accuracy when testing on VGG11 and protecting VGG16 with different $N$ settings.}
    \label{fig:N}
\end{figure}

Then, we set $\alpha=4$ and test different $N$ values as shown in Tab.~\ref{tab:N}. Fig.~\ref{fig:N} is a plot of the accuracy when protecting VGG16 and considering VGG11 as the target model. When $N>100$, the performance drop of the target model becomes slower with increasing $N$. Thus, in Sec.~\ref{sec:exp}, we use $N = 100$ when generating the privacy-protected images.
\begin{table}[h!]
\centering
\resizebox{0.85\columnwidth}{!}{
\begin{tabular}{|c|c|c|c|c|c|c|}\hline
  Accuracy (\%)                & VGG11 & VGG16 & ResNet18 & ResNet34 & W\_Res50\_2 & Mob\_v2\\\hline
$N=10$                   & 25.6                      & 100                       & 28.9                         & 37.8                         & 23.6                            & 24.4                        \\\hline 
$N=20$                   & 14.5                      & 100                       & 16.8                         & 28.3                         & 16.2                            & 16.1                        \\\hline 
$N=50$                   & 9.5                       & 99.8                      & 9.7                          & 19.9                         & 11.8                            & 11.1                        \\\hline 
$N=100$                  & 8                         & 99.8                      & 8.5                          & 16.9                         & 10.4                            & 9.8                         \\\hline 
$N=200$                  & 6.8                       & 99.8                      & 7.6                          & 16.2                         & 10.3                            & 10.9       \\\hline                
\end{tabular}}
\vspace{-0.2cm}
\caption{Results for protecting VGG16 under different $N$ settings.}
\label{tab:N}
\vspace{-0.6cm}
\end{table}

\subsection{Image Quality and Protection Performance}
\label{sec:quality}
\begin{figure}[ht!]
    \centering\resizebox{0.8\columnwidth}{!}{
    \includegraphics{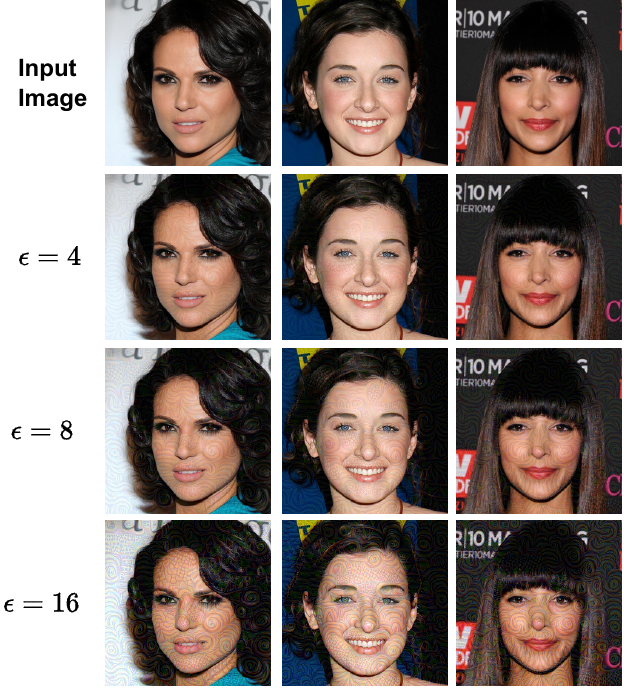}}
    \vspace{-0.2cm}
    \caption{Example generated images for different $\epsilon$ values.}
    \label{fig:quality}
    \vspace{-0.2cm}
\end{figure}

We use a budget of $\epsilon$ to control how much generated image can differ from the original. Thus, $\epsilon$ affects the generated image quality. We perform experiments on the Celeba-HQ dataset and protect VGG16 to analyze the trade-off between the image quality and protection performance. The image quality is measured by the Structural Similarity Index Measure (SSIM) between the original and generated image. The range of SSIM is 0-1, the higher being the better. The results in Tab.~\ref{tab:quality} show that as $\epsilon$ decreases, the image quality increases, but the degradation in the performance of attacked models decreases, as expected. Example images in Fig.~\ref{fig:quality} show the visual quality when using different values of $\epsilon$.

\begin{table}[ht!]
\resizebox{1\columnwidth}{!}{
\begin{tabular}{|c|c|c|c|c|c|c|c|}\hline
Accuracy (\%)                & VGG11 & VGG16 & ResNet18 & ResNet34 & W-Res50-2 & Mob-v2 &SSIM\\\hline
$\epsilon=4$                    & 86.56                     & 100                       & 52.19                        & 55.94                        & 49.84                                 & 39.68                             & 0.973                    \\\hline
$\epsilon=8$                    & 25.31                     & 100                       & 13.13                        & 18.44                        & 17.5                                  & 9.06                              & 0.909                    \\\hline
$\epsilon=16$                   & 2.37                      & 99.53                     & 4.73                         & 6.78                         & 7.73                                  & 1.1                               & 0.754   \\\hline                
\end{tabular}}
\vspace{-0.2cm}
\caption{Image quality results, when protecting VGG16, using different $\epsilon$ values. Image quality is measured with SSIM.}
\label{tab:quality}
\vspace{-0.2cm}
\end{table}
\begin{table}[h!]
\resizebox{1\columnwidth}{!}{
\begin{tabular}{|c|c|c|c|c|c|c|}\hline
Accuracy (\%)                & VGG11 & VGG16 & ResNet18 & ResNet34 & W-Res50-2 & Mob-v2 \\\hline
VGG11             & 100                       & 19.23                     & 11.54                        & 42.31                        & 15.38                                 & 15.38                             \\\hline
VGG16             & 23.08                     & 100                       & 11.54                        & 30.77                        & 11.54                                 & 19.23                             \\\hline
ResNet18          & 42.31                     & 53.85                     & 100                          & 19.23                        & 26.92                                 & 23.08                             \\\hline
ResNet34          & 61.54                     & 50                        & 15.38                        & 100                          & 30.77                                 & 30.77                             \\\hline
W-Res50-2 & 57.69                     & 34.62                     & 7.69                         & 30.77                        & 100                                   & 34.62                             \\\hline
Mob-v2     & 73.08                     & 30.77                     & 15.38                        & 57.69                        & 50                                    & 100           \\\hline                   
\end{tabular}}
\vspace{-0.2cm}
\caption{Results on the Celeba-HQ dataset using 10 classes.}
\label{tab:10c}
\end{table}
\begin{table}[h!]
\resizebox{1\columnwidth}{!}{
\begin{tabular}{|c|c|c|c|c|c|c|}\hline
Accuracy (\%)                & VGG11 & VGG16 & ResNet18 & ResNet34 & W-Res50-2 & Mob-v2 \\\hline
VGG11             & 100                       & 4.08                      & 3.06                         & 5.1                          & 9.18                                  & 5.1                               \\\hline
VGG16             & 6.12                      & 100                       & 1.02                         & 1.02                         & 5.1                                   & 4.08                              \\\hline
ResNet18          & 16.33                     & 5.1                       & 100                          & 7.14                         & 14.29                                 & 7.14                              \\\hline
ResNet34          & 19.39                     & 4.08                      & 2.04                         & 100                          & 15.31                                 & 8.16                              \\\hline
W-Res50-2 & 26.53                     & 6.12                      & 3.06                         & 4.08                         & 100                                   & 2.04                              \\\hline
Mob-v2     & 35.71                     & 7.14                      & 15.31                        & 15.31                        & 19.39                                 & 100             \\\hline                 
\end{tabular}}
\vspace{-0.2cm}
\caption{Results on Celeba-HQ dataset using 50 classes.}
\label{tab:50c}
\vspace{-0.3cm}
\end{table}
\subsection{Performance Analysis of Task Complexity}
\label{sec:task_comp}

In this study, we reduce the task complexity by reducing the number of classes on the Celeba-HQ dataset. For the models, we only change the final layer of the MLP to match the number of classes. The results for 10 and 50 classes are shown in Tab.~\ref{tab:10c} and Tab.~\ref{tab:50c}, respectively. Compared to the results from 307 classes in Tab.~\ref{tab:celeba}, the average performance drop on target models decreased from 93.7\% to 68.46\% and 90.75\% for 10 classes and 50 classes, respectively. From these results, we can deduce that the effectiveness of our approach increases as the complexity of the task increases.

%% file: sec/8_conclusion.tex
\section{Conclusion}
We have introduced a novel approach for safeguarding authorized models and deceiving unauthorized black-box models, addressing practical privacy concerns with real-world use cases. Our method excels in preserving privacy for the classification tasks involving a relatively large number of classes, achieving strong protection accuracy of $100$\% (ImageNet), $99.92$\% (Celeba-HQ), and $99.67$\%(AffectNet), while significantly reducing unauthorized model accuracy, on average to $11.97$\%, $6.63$\%, and $55.5$\%, respectively, for six different attacked models. Our in-depth evaluation includes the investigation of the underlying reasoning of the proposed FMD-based attack’s efficacy, verification of a positive correlation between the number of classes and the proposed approach’s effectiveness, attack transferability across tasks to unknown models (yet, such transferability is affected by the task and model characteristics), and a comprehensive ablation study. Future work includes developing an approach that can maintain effectiveness, regardless of the complexity of the tasks, with better cross-task transferability.

%% file: ijcai24.bbl
\begin{thebibliography}{}

\bibitem[\protect\citeauthoryear{Carlini and Wagner}{2017}]{carlini2017towards}
Nicholas Carlini and David Wagner.
\newblock Towards evaluating the robustness of neural networks.
\newblock In {\em 2017 ieee symposium on security and privacy (sp)}, pages 39--57. Ieee, 2017.

\bibitem[\protect\citeauthoryear{Chakraborty \bgroup \em et al.\egroup }{2018}]{chakraborty2018adversarial}
Anirban Chakraborty, Manaar Alam, Vishal Dey, Anupam Chattopadhyay, and Debdeep Mukhopadhyay.
\newblock Adversarial attacks and defences: A survey.
\newblock {\em arXiv preprint arXiv:1810.00069}, 2018.

\bibitem[\protect\citeauthoryear{Dang \bgroup \em et al.\egroup }{2017}]{dang2017evading}
Hung Dang, Yue Huang, and Ee-Chien Chang.
\newblock Evading classifiers by morphing in the dark.
\newblock In {\em Proceedings of the 2017 ACM SIGSAC conference on computer and communications security}, pages 119--133, 2017.

\bibitem[\protect\citeauthoryear{De~Wolf}{2020}]{de2020contextualizing}
Ralf De~Wolf.
\newblock Contextualizing how teens manage personal and interpersonal privacy on social media.
\newblock {\em New media \& society}, 22(6):1058--1075, 2020.

\bibitem[\protect\citeauthoryear{Deng \bgroup \em et al.\egroup }{2009}]{deng2009imagenet}
Jia Deng, Wei Dong, Richard Socher, Li-Jia Li, Kai Li, and Li~Fei-Fei.
\newblock Imagenet: A large-scale hierarchical image database.
\newblock In {\em 2009 IEEE Conference on Computer Vision and Pattern Recognition}, pages 248--255. IEEE, 2009.

\bibitem[\protect\citeauthoryear{Dosovitskiy \bgroup \em et al.\egroup }{2014}]{dosovitskiy2014discriminative}
Alexey Dosovitskiy, Jost~Tobias Springenberg, Martin Riedmiller, and Thomas Brox.
\newblock Discriminative unsupervised feature learning with convolutional neural networks.
\newblock {\em Advances in neural information processing systems}, 27, 2014.

\bibitem[\protect\citeauthoryear{Fei \bgroup \em et al.\egroup }{2020}]{fei2020deep}
Zixiang Fei, Erfu Yang, David Day-Uei Li, Stephen Butler, Winifred Ijomah, Xia Li, and Huiyu Zhou.
\newblock Deep convolution network based emotion analysis towards mental health care.
\newblock {\em Neurocomputing}, 388:212--227, 2020.

\bibitem[\protect\citeauthoryear{Goodfellow \bgroup \em et al.\egroup }{2014}]{goodfellow2014explaining}
Ian~J Goodfellow, Jonathon Shlens, and Christian Szegedy.
\newblock Explaining and harnessing adversarial examples.
\newblock {\em arXiv preprint arXiv:1412.6572}, 2014.

\bibitem[\protect\citeauthoryear{Guo \bgroup \em et al.\egroup }{2017}]{guo2017calibration}
Chuan Guo, Geoff Pleiss, Yu~Sun, and Kilian~Q Weinberger.
\newblock On calibration of modern neural networks.
\newblock In {\em International conference on machine learning}, pages 1321--1330. PMLR, 2017.

\bibitem[\protect\citeauthoryear{Guo \bgroup \em et al.\egroup }{2019}]{guo2019simple}
Chuan Guo, Jacob Gardner, Yurong You, Andrew~Gordon Wilson, and Kilian Weinberger.
\newblock Simple black-box adversarial attacks.
\newblock In {\em International Conference on Machine Learning}, pages 2484--2493. PMLR, 2019.

\bibitem[\protect\citeauthoryear{He \bgroup \em et al.\egroup }{2016}]{he2016resnet}
Kaiming He, Xiangyu Zhang, Shaoqing Ren, and Jian Sun.
\newblock Deep residual learning for image recognition.
\newblock In {\em Proceedings of the IEEE conference on computer vision and pattern recognition}, pages 770--778, 2016.

\bibitem[\protect\citeauthoryear{Huang \bgroup \em et al.\egroup }{2019}]{huang2019toward}
Zhengbai Huang, Meng Zhang, and Yi~Zhang.
\newblock Toward efficient encrypted image retrieval in cloud environment.
\newblock {\em IEEE Access}, 7:174541--174550, 2019.

\bibitem[\protect\citeauthoryear{Ilyas \bgroup \em et al.\egroup }{2018}]{ilyas2018black}
Andrew Ilyas, Logan Engstrom, Anish Athalye, and Jessy Lin.
\newblock Black-box adversarial attacks with limited queries and information.
\newblock In {\em International conference on machine learning}, pages 2137--2146. PMLR, 2018.

\bibitem[\protect\citeauthoryear{Juuti \bgroup \em et al.\egroup }{2019}]{juuti2019making}
Mika Juuti, Buse Gul~Atli, and N~Asokan.
\newblock Making targeted black-box evasion attacks effective and efficient.
\newblock In {\em Proceedings of the 12th ACM Workshop on Artificial Intelligence and Security}, pages 83--94, 2019.

\bibitem[\protect\citeauthoryear{Kim and Kim}{2017}]{kim2017convolutional}
Phil Kim and Phil Kim.
\newblock Convolutional neural network.
\newblock {\em MATLAB deep learning: with machine learning, neural networks and artificial intelligence}, pages 121--147, 2017.

\bibitem[\protect\citeauthoryear{Ko \bgroup \em et al.\egroup }{2023}]{ko2023multi}
Kyoungmin Ko, SungHwan Kim, and Hyun Kwon.
\newblock Multi-targeted audio adversarial example for use against speech recognition systems.
\newblock {\em Computers \& Security}, 128:103168, 2023.

\bibitem[\protect\citeauthoryear{Kwon \bgroup \em et al.\egroup }{2018}]{kwon2018friend}
Hyun Kwon, Yongchul Kim, Ki-Woong Park, Hyunsoo Yoon, and Daeseon Choi.
\newblock Friend-safe evasion attack: An adversarial example that is correctly recognized by a friendly classifier.
\newblock {\em computers \& security}, 78:380--397, 2018.

\bibitem[\protect\citeauthoryear{Kwon \bgroup \em et al.\egroup }{2019a}]{kwon2019selective}
Hyun Kwon, Yongchul Kim, Hyunsoo Yoon, and Daeseon Choi.
\newblock Selective audio adversarial example in evasion attack on speech recognition system.
\newblock {\em IEEE Transactions on Information Forensics and Security}, 15:526--538, 2019.

\bibitem[\protect\citeauthoryear{Kwon \bgroup \em et al.\egroup }{2019b}]{kwon2019restricted}
Hyun Kwon, Hyunsoo Yoon, and Daeseon Choi.
\newblock Restricted evasion attack: Generation of restricted-area adversarial example.
\newblock {\em IEEE Access}, 7:60908--60919, 2019.

\bibitem[\protect\citeauthoryear{Kwon \bgroup \em et al.\egroup }{2022}]{kwon2022priority}
Hyun Kwon, Changhyun Cho, and Jun Lee.
\newblock Priority evasion attack: An adversarial example that considers the priority of attack on each classifier.
\newblock {\em IEICE TRANSACTIONS on Information and Systems}, 105(11):1880--1889, 2022.

\bibitem[\protect\citeauthoryear{Kwon}{2021}]{kwon2021dual}
Hyun Kwon.
\newblock Dual-targeted textfooler attack on text classification systems.
\newblock {\em IEEE Access}, 2021.

\bibitem[\protect\citeauthoryear{Kwon}{2023}]{kwon2023toward}
Hyun Kwon.
\newblock Toward selective adversarial attack for gait recognition systems based on deep neural network.
\newblock {\em IEICE TRANSACTIONS on Information and Systems}, 106(2):262--266, 2023.

\bibitem[\protect\citeauthoryear{Li \bgroup \em et al.\egroup }{2021}]{li2021framework}
Ning Li, Liang Cheng, Lingyong Huang, Chen Ji, Min Jing, Zhixin Duan, Jingjing Li, and Manchun Li.
\newblock Framework for unknown airport detection in broad areas supported by deep learning and geographic analysis.
\newblock {\em IEEE Journal of Selected Topics in Applied Earth Observations and Remote Sensing}, 14:6328--6338, 2021.

\bibitem[\protect\citeauthoryear{Lin \bgroup \em et al.\egroup }{2014}]{lin2014microsoft}
Tsung-Yi Lin, Michael Maire, Serge Belongie, James Hays, Pietro Perona, Deva Ramanan, Piotr Doll{\'a}r, and C~Lawrence Zitnick.
\newblock Microsoft coco: Common objects in context.
\newblock In {\em Computer Vision--ECCV 2014: 13th European Conference, Zurich, Switzerland, September 6-12, 2014, Proceedings, Part V 13}, pages 740--755. Springer, 2014.

\bibitem[\protect\citeauthoryear{Lin \bgroup \em et al.\egroup }{2017}]{lin2017focal}
Tsung-Yi Lin, Priya Goyal, Ross Girshick, Kaiming He, and Piotr Doll{\'a}r.
\newblock Focal loss for dense object detection.
\newblock In {\em Proceedings of the IEEE International Conference on Computer Vision (ICCV)}, pages 2980--2988. IEEE, 2017.

\bibitem[\protect\citeauthoryear{Liu \bgroup \em et al.\egroup }{2015}]{liu2015celebahq}
Ziwei Liu, Ping Luo, Xiaogang Wang, and Xiaoou Tang.
\newblock Celeba-hq: A high-quality dataset of celebrity images.
\newblock {\em arXiv preprint arXiv:1710.10196}, 2015.

\bibitem[\protect\citeauthoryear{Liu \bgroup \em et al.\egroup }{2016}]{liu2016ssd}
Wei Liu, Dragomir Anguelov, Dumitru Erhan, Christian Szegedy, Scott Reed, Cheng-Yang Fu, and Alexander~C. Berg.
\newblock Ssd: Single shot multibox detector.
\newblock In {\em European conference on computer vision}, pages 21--37. Springer, 2016.

\bibitem[\protect\citeauthoryear{Maho \bgroup \em et al.\egroup }{2021}]{maho2021surfree}
Thibault Maho, Teddy Furon, and Erwan Le~Merrer.
\newblock Surfree: a fast surrogate-free black-box attack.
\newblock In {\em Proceedings of the IEEE/CVF Conference on Computer Vision and Pattern Recognition}, pages 10430--10439, 2021.

\bibitem[\protect\citeauthoryear{Marcel and others}{2023}]{torchvision}
Adam Marcel et~al.
\newblock torchvision: Datasets, transforms and models specific to computer vision.
\newblock \url{https://pytorch.org/vision/}, 2023.
\newblock Accessed: Jan 17 2024.

\bibitem[\protect\citeauthoryear{Mollahosseini \bgroup \em et al.\egroup }{2017}]{mollahosseini2017affectnet}
Ali Mollahosseini, Behzad Hasani, and Mohammad~H Mahoor.
\newblock Affectnet: A database for facial expression, valence, and arousal computing in the wild.
\newblock {\em IEEE Transactions on Affective Computing}, 10(1):18--31, 2017.

\bibitem[\protect\citeauthoryear{Nicol Turner~Lee}{2022}]{police}
Caitlin Chin-Rothmann Nicol Turner~Lee.
\newblock Police surveillance and facial recognition: Why data privacy is imperative for communities of color, 2022.
\newblock Supplied as supplemental material {\tt tr.pdf}.

\bibitem[\protect\citeauthoryear{Papernot \bgroup \em et al.\egroup }{2017}]{papernot2017practical}
Nicolas Papernot, Patrick McDaniel, Ian Goodfellow, Somesh Jha, Z~Berkay Celik, and Ananthram Swami.
\newblock Practical black-box attacks against machine learning.
\newblock In {\em Proceedings of the 2017 ACM on Asia conference on computer and communications security}, pages 506--519, 2017.

\bibitem[\protect\citeauthoryear{Ren \bgroup \em et al.\egroup }{2015}]{ren2015faster}
Shaoqing Ren, Kaiming He, Ross Girshick, and Jian Sun.
\newblock Faster r-cnn: Towards real-time object detection with region proposal networks.
\newblock In {\em Advances in neural information processing systems}, 2015.

\bibitem[\protect\citeauthoryear{Sandler \bgroup \em et al.\egroup }{2018}]{sandler2018mobilenetv2}
Mark Sandler, Andrew Howard, Menglong Zhu, Andrey Zhmoginov, and Liang-Chieh Chen.
\newblock Mobilenetv2: Inverted residuals and linear bottlenecks.
\newblock In {\em Proceedings of the IEEE conference on computer vision and pattern recognition}, pages 4510--4520, 2018.

\bibitem[\protect\citeauthoryear{Selvaraju \bgroup \em et al.\egroup }{2016}]{selvaraju2016grad}
Ramprasaath~R Selvaraju, Abhishek Das, Ramakrishna Vedantam, Michael Cogswell, Devi Parikh, and Dhruv Batra.
\newblock Grad-cam: Why did you say that?
\newblock {\em arXiv preprint arXiv:1611.07450}, 2016.

\bibitem[\protect\citeauthoryear{Serengil and Ozpinar}{2020}]{serengil2020lightface}
Sefik~Ilkin Serengil and Alper Ozpinar.
\newblock Lightface: A hybrid deep face recognition framework.
\newblock In {\em 2020 innovations in intelligent systems and applications conference (ASYU)}, pages 1--5. IEEE, 2020.

\bibitem[\protect\citeauthoryear{Simonyan and Zisserman}{2015}]{simonyan2014vgg}
Karen Simonyan and Andrew Zisserman.
\newblock Very deep convolutional networks for large-scale image recognition.
\newblock In {\em International Conference on Learning Representations}, 2015.

\bibitem[\protect\citeauthoryear{Trepte}{2021}]{trepte2021social}
Sabine Trepte.
\newblock The social media privacy model: Privacy and communication in the light of social media affordances.
\newblock {\em Communication Theory}, 31(4):549--570, 2021.

\bibitem[\protect\citeauthoryear{Vlontzos \bgroup \em et al.\egroup }{2019}]{vlontzos2019multiple}
Athanasios Vlontzos, Amir Alansary, Konstantinos Kamnitsas, Daniel Rueckert, and Bernhard Kainz.
\newblock Multiple landmark detection using multi-agent reinforcement learning.
\newblock In {\em Medical Image Computing and Computer Assisted Intervention--MICCAI 2019: 22nd International Conference, Shenzhen, China, October 13--17, 2019, Proceedings, Part IV 22}, pages 262--270. Springer, 2019.

\bibitem[\protect\citeauthoryear{Wang and Kosinski}{2018}]{wang2018deep}
Yilun Wang and Michal Kosinski.
\newblock Deep neural networks are more accurate than humans at detecting sexual orientation from facial images.
\newblock {\em Journal of personality and social psychology}, 114(2):246, 2018.

\bibitem[\protect\citeauthoryear{Wang \bgroup \em et al.\egroup }{2020}]{wang2020cnn}
Zijie~J Wang, Robert Turko, Omar Shaikh, Haekyu Park, Nilaksh Das, Fred Hohman, Minsuk Kahng, and Duen Horng~Polo Chau.
\newblock Cnn explainer: learning convolutional neural networks with interactive visualization.
\newblock {\em IEEE Transactions on Visualization and Computer Graphics}, 27(2):1396--1406, 2020.

\bibitem[\protect\citeauthoryear{Wang \bgroup \em et al.\egroup }{2021}]{wang2021convolutional}
Zi~Wang, Chengcheng Li, and Xiangyang Wang.
\newblock Convolutional neural network pruning with structural redundancy reduction.
\newblock In {\em Proceedings of the IEEE/CVF conference on computer vision and pattern recognition}, pages 14913--14922, 2021.

\bibitem[\protect\citeauthoryear{Wang}{2022}]{wang2022presentation}
Dawei Wang.
\newblock Presentation in self-posted facial images can expose sexual orientation: Implications for research and privacy.
\newblock {\em Journal of Personality and Social Psychology}, 122(5):806, 2022.

\bibitem[\protect\citeauthoryear{Xu \bgroup \em et al.\egroup }{2011}]{xu2011information}
Heng Xu, Tamara Dinev, Jeff Smith, and Paul Hart.
\newblock Information privacy concerns: Linking individual perceptions with institutional privacy assurances.
\newblock {\em Journal of the Association for Information Systems}, 12(12):1, 2011.

\bibitem[\protect\citeauthoryear{Zagoruyko and Komodakis}{2016}]{zagoruyko2016wide}
Sergey Zagoruyko and Nikos Komodakis.
\newblock Wide residual networks.
\newblock In {\em British Machine Vision Conference}, 2016.

\end{thebibliography}
